\documentclass[letterpaper, 10 pt, conference]{ieeeconf}  

\IEEEoverridecommandlockouts                              

\usepackage[tracking=true]{microtype}
\usepackage{amsmath} 
\usepackage{amssymb}  
\usepackage{graphicx}
\usepackage{epsfig} 
\usepackage{mathptmx} 
\usepackage{times} 
\usepackage{amsmath} 
\usepackage{amssymb}  
\usepackage{xspace}
\usepackage[font=small]{caption}
\usepackage[table]{xcolor}

\usepackage[frozencache,cachedir=minted]{minted}
\usepackage{hyperref}
\setminted{fontsize=\footnotesize}
\usepackage{url}
\usepackage[]{xcolor}
\usepackage{caption}
\usepackage{booktabs} 
\usepackage{tikz}
\usetikzlibrary{positioning,shapes.geometric,arrows,fit,shapes.symbols,svg.path,tikzmark,calc}
\usepackage{booktabs}
\usepackage{textcomp}
\usepackage{listings}
\usepackage{subcaption}
\pgfdeclarelayer{background}
\pgfdeclarelayer{foreground}
\pgfsetlayers{background,main,foreground}
\usepackage{siunitx}
\usepackage{adjustbox}
\usepackage{array}
\usepackage{multirow}

\let\labelindent\relax 

\usepackage[inline]{enumitem} 

\newcommand{\algname}{FogROS2-Config\xspace}

\title{\LARGE \bf
FogROS2-Config: A Toolkit for Choosing Server Configurations for Cloud Robotics
}

\author{ Kaiyuan Chen$^{*1}$, Kush Hari$^{*1A}$, Rohil Khare$^{1A}$, Charlotte Le$^{1A}$,  
Trinity Chung$^{1A}$,  \\
Jaimyn Drake$^{1A}$,
Jeffrey Ichnowski$^{3}$,
 John Kubiatowicz$^{1}$, 
 and Ken Goldberg$^{1,2,A}$
\thanks{$^{*}$Equal Contribution}
\thanks{$^{1}$Department of Electrical Engineering and Computer Sciences}%
\thanks{$^{A}$The AUTOLab at UC Berkeley (\href{http://automation.berkeley.edu}{automation.berkeley.edu}).}
\thanks{$^{2}$Department of Industrial Engineering and Operations Research}%
\thanks{$^{1,2}$University of California, Berkeley, CA, USA }%
\thanks{$^{3}$Robotics Institute, Carnegie Mellon University}%
}

\begin{document}

\maketitle
\thispagestyle{empty}
\pagestyle{empty}

\begin{abstract}
Cloud service providers provide over 50,000 distinct and dynamically changing set of cloud server options. To help roboticists make cost-effective decisions, we present \algname, an open toolkit that takes ROS2 nodes as input and automatically runs relevant benchmarks to quickly return a menu of cloud compute services that tradeoff latency and cost. 
Because it is infeasible to try every hardware configuration, \algname quickly samples tests a small set of edge-case servers. 
We evaluate \algname on three robotics application tasks: visual SLAM, grasp planning. and motion planning. \algname can reduce the cost by up to 20x.
By comparing with a Pareto frontier for cost and latency by running the application task on feasible server configurations, 
we evaluate cost and latency models and confirm that \algname selects efficient hardware configurations to balance cost and latency. 
Videos and code are available on the website 
\url{https://sites.google.com/view/fogros2-config}
\end{abstract}



\section{Introduction}

Many new robotics applications require powerful computational resources (such as GPUs, FPGAs, and TPUs), making it impractical and uneconomical to deploy on onboard robot hardware. 
Recently, the emergence of cloud robotics allows robots to operate with more affordable onboard compute hardware by leveraging a cloud-based Software as a Service (SaaS) model of computing. Robots can access and pay for compute resources as needed, to reduce the cost of deployment and operation. For example, an average lifespan of a home vacuum robot is 6 years and it runs 20 minutes per week. If we run the same task on a much faster AWS cloud machine (t2.medium), it takes only \$6.00 for 6 years. In contrast, a typical single-board computer (such as Raspberry Pi), with hardware and energy cost more than \$100.00.

While there are clear cost advantages in using the cloud, the performance-cost trade-off is understudied in cloud robotics due to the following four challenges: 
(1) Cloud service providers offer many different interfaces and pricing structures, presupposing that users already understand the provider-specific machine type required for their needs, 
(2) Providers typically employ coarse-grained monthly or hourly rates but update prices hourly.
(3) The trade-off of application latency and operating cost is often unknown or variable: slower machine execution might lead to prolonged machine usage and potentially higher costs,
(4) Robotics experts often know the high-level application needs but may lack specific domain knowledge about the optimal compute specifications and cloud machine types for those requirements.

\begin{figure}
    \centering
    \includegraphics[width=\linewidth]{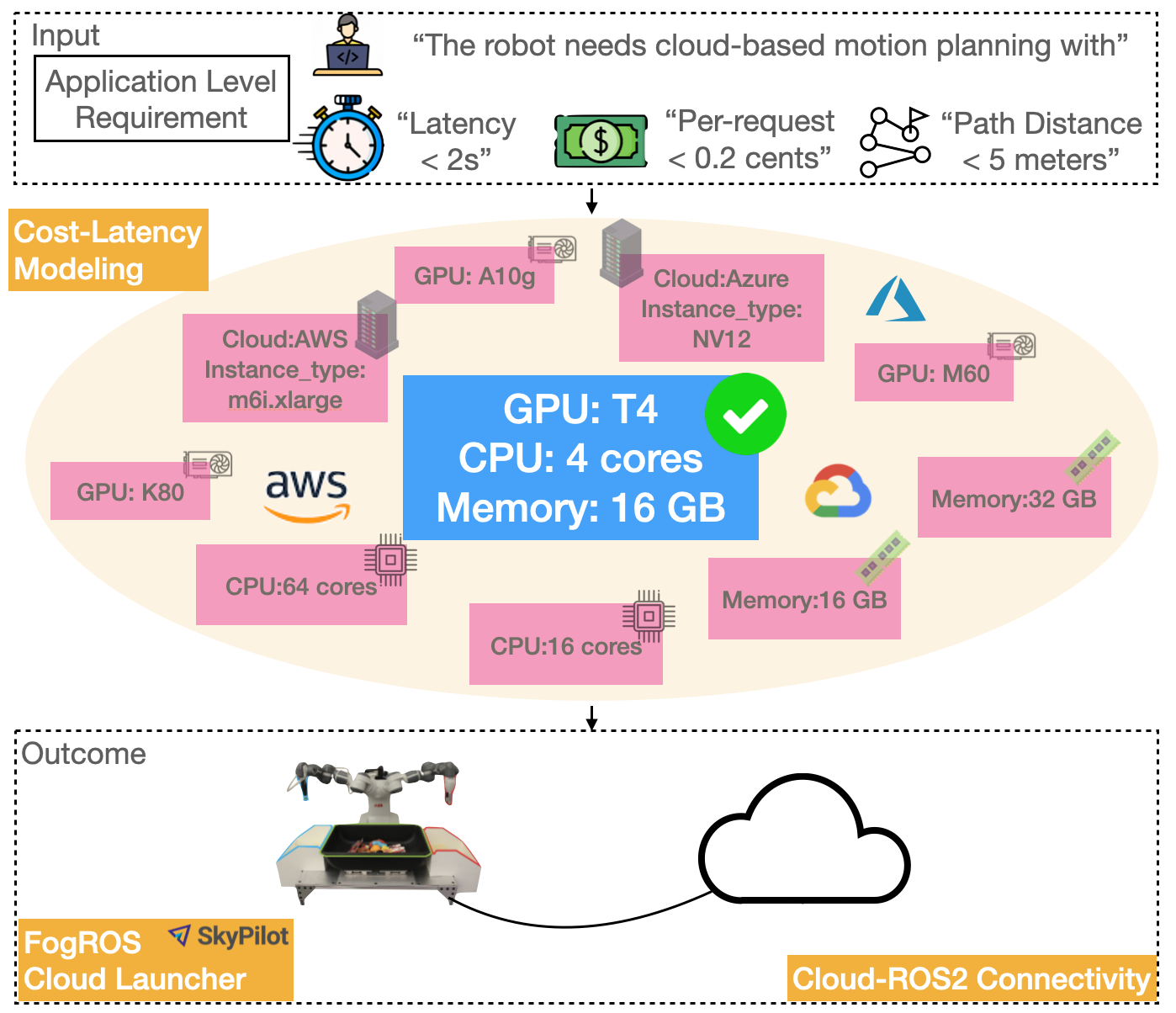}
    
    \caption{\textbf{A Sample Use Case of \algname.} With \algname, users only need to input application-level requirements, such as latency and per-request cost. \algname automates the cloud machine selection by modeling the latency cost tradeoff and facilities the cost-effective cloud robotics machine selection. \algname automatically provisions the cloud machines and enables unmodified ROS2 applications to run as if all components are on the local robot.
    }
    \label{fig:enter-label}
\end{figure}

We present \algname, an openly available cloud robotics platform that automatically models cost and performance trade-offs for unmodified ROS2 applications at a per-request granularity. 
\algname automates the latency analysis using Skypilot~\cite{286502}, an inter-cloud broker that orchestrates heterogeneous providers of cloud services.
With over 50,000 available selections of cloud instances, it is infeasible to model the cost and latency for each instance to compute the corresponding optimum. However, \algname models cost and latency by running benchmark tests on a small set of edge-case servers.
\algname models the cost and latency trade-off of how application performance corresponds to different servers and how that is mapped to specific machine types with various cloud service providers.
Directly based on the application-level requirements, \algname helps roboticists choose the cloud service provider and hardware specification. 

We evaluate \algname by running it on three cloud robotics applications: visual Simultaneous Localization and Mapping (vSLAM), motion planning, and grasp planning. For each application, we compare the model results to a ground-truth Pareto frontier formed by running the application on possible server configurations. We show that \algname can reduce the benchmark cost by up to 20 times.

\begin{figure*}
    \centering
    \includegraphics[width=0.75\linewidth]{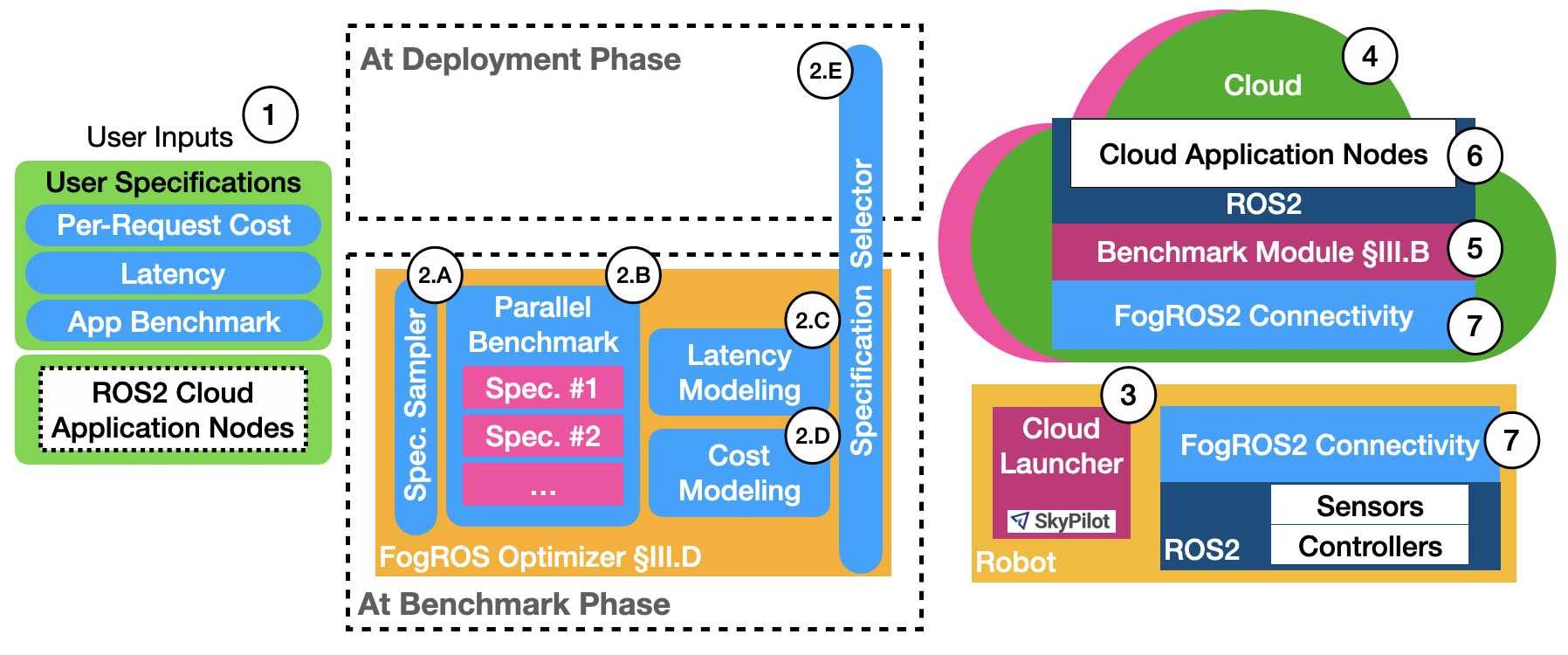}
    \caption{\textbf{An Overview of the Benchmark and Launch Sequence of \algname} \algname automatically computes the most cost-effective cloud machine configuration through its optimizer.  \algname's model can be reused with different cost and latency constraints without the need to rerun the benchmark.
    The launch sequence is elaborated in Section \ref{sec:design:launch} with the same sequence numbers. (1) User specifies the application level constraints; (2) \algname finds the optimal hardware specification; (3-6) \algname launches the cloud server, provisions the ROS2 environment, and offloads the application nodes (7) \algname connects the robot to the cloud server.}
    \label{fig:overview} 
\end{figure*}
%

The paper makes the following contributions: 
(1) \algname, an open-source cloud robotics platform that efficiently computes a fine-grained cost analysis of ROS2 applications, 
(2) An algorithm that takes user-specified latency and cost requirements and determines the cost-effective cloud instance for a robotic application based on the user's specifications,
(3) Data from experiments on three robotics applications with distinct cost-performance trade-offs.

\section{Related Work}


Since James Kuffner introduced the term ‘Cloud Robotics' in 2010, there has been significant progress in cloud and fog computing. This has been applied to cloud robotics for a variety of applications \cite{kehoe2015survey} such as multi-robot SLAM \cite{huang_edge_2022}, cloth-folding \cite{hoque_learning_2022}, semantic inventory monitoring \cite{Rashid24lifelonglerf}, autonomous vehicles \cite{schafhalter_leveraging_2023}, path-planning \cite{lam2014path, chen_cloud-based_2021}, grasping \cite{kehoe2012toward}, and grasp planning \cite{li2018dex}. 
In tandem, cloud services, such as Amazon AWS, Google Cloud Platform, and Microsoft
Azure, have matured and developed to now offer a more diverse range of computing options and tools \cite{goudarzi_cloud_2022}. This makes them more attractive, especially as roboticists seek to employ more computationally intensive machine learning approaches, such as neural networks and transformers, in their designs.
Also, ROS2 has become the de-facto platform for building robotics applications by breaking down the application into standalone nodes that interact with each other using the pub/sub paradigm. The FogROS body of work, including FogROS \cite{chen2021fogros}, FogROS2 \cite{ichnowski2022fogros}, and FogROS2-SGC~\cite{chen_fogros2-sgc_2023}, seeks to make it easier for roboticists to use increasingly complex yet powerful cloud resources using ROS/ROS2 across different robots and cloud platforms in a secure way. FogROS2-LS~\cite{chen2024fogrosls} focuses on the network routing of latency-sensitive applications and complements the present paper. 



FogROS2 \cite{ichnowski2022fogros} is the first cloud robotics platform that supports multiple major cloud service providers, including Amazon AWS, Google Cloud Platform, and Microsoft Azure, allowing users to access them from within the ROS2 ecosystem. 
Sky Computing \cite{10.1145/3458336.3465301, chasins_sky_2022} is a future view of cloud computing that seeks to facilitate spreading compute across different cloud compute providers. 
\algname embraces the synergy of its multi-cloud paradigm by extending SkyPilot~\cite{286502}, a framework that handles the deployment and execution of a user's job on as many cloud providers as required. The Experiment Management System (EMS) for Robotics ~\cite{lin2023ems} proposed a massive computational experiment management for robotics tasks, which uses Sky for the deployment of robotics experiments on the heterogeneous cloud environment.  
We recognize that both EMS~\cite{lin2023ems} and SkyPilot~\cite{286502} focus on model training and require extensive effort to adapt to general robotics applications, which are not necessarily learning-based models.
In this paper, we query SkyPilot for the cheapest hardware given the user specification. Combined with the \algname, robots can seamlessly launch  cloud instances by specifying application level requirements. 

\section{\algname Problem Formulation and Features}

\subsection{Problem Formulation}
\algname helps robotocists decide how a robot can offload part of its computational graph in ROS2 to a cloud machine. Given the user-defined constraints of time-per-request and cost-per-request on each partition, it models cost and latency functions to approximate the best available hardware specification that fulfills the constraint or reports that such a specification does not exist. 
Time-per-request is defined as the computational latency performed on the server, which excludes data transfer time that specific machine type used has a limited effect on data transfer.
We formulate this approximation as a relaxed integer optimization problem that identifies cloud resources (CPU count, GPU type, and memory) that align with a user-provided task, cost and latency constraints. 

\textbf{Assumptions} 
We assume the robotic algorithm is iterative and its timing is deterministic or stochastic with low variance. We also assume that CPU and memory values for the hardware configuration are discrete and only exist in predefined sets as setup in cloud service providers. Therefore, we perform a relaxation on the optimization problem by creating a solution subset of the nearest feasible options. 

\subsection{\algname System Features}

\textbf{Minimal ROS2 Application Modification}.
\algname adheres to the abstraction of ROS2 and offloads unmodified ROS2 applications to the cloud, 
but they work as if all of the ROS2 nodes were on the same machine. 

\textbf{Unified Interface for Heterogeneous Cloud Providers}.
\algname provides a unified interface to interact with multiple cloud service providers. One can use standard ROS2 launch file interfaces without relying on cloud service provider dependent interfaces. 

\textbf{High Level Cloud Specification Description}.
In traditional approaches, selecting hardware specifications involves empirical methods, such as manual web browsing, often leading to sub-optimal performance and cost.
In contrast, \algname allows users to directly input application-level requirements, such as per-request cost and latency. \algname automates a relaxed cost optimization to identify the most cost-effective hardware and its corresponding cloud machine type for a given cloud service provider. 
 


\subsection{\algname Workflow } 
\label{sec:design:launch}

Figure \ref{fig:overview} shows the sequence diagram of \algname.
The following are the steps of \algname launcher that launches the cloud instances:
(1) Users provide the ROS2 nodes that need to be offloaded to the cloud and specify the latency and per-request cost constraints of the ROS2 nodes, 
(2) Execute the \algname optimizer to find the optimal cloud machine specification,
(3) Use Skypilot to launch and manage cloud virtual machines, 
(4) Provision the cloud virtual machines by setting up the ROS2 environment and workspace,
(5) Setup the benchmark module for monitoring the latency and cost performance of the application,
(6) Launch the application ROS2 nodes on the cloud,
(7) Launch FogROS2 cloud connectivity to securely and globally connect the robot with the multi-cloud virtual machines.
At the same time, the ROS2 launch system initiates ROS2 nodes on the robot.

For first time user, the \algname optimizer at step (2) finds the cost-effective hardware that fulfills the constraint by sampling (2.A) and benchmarking (2.B) across various hardware specifications, as well as modeling the latency (2.C) and cost (2.D). 
The modeled latency and cost can be used for future to select the optimal hardware specification with different user constraints and requirements (2.E).

\section{\algname Design}


\algname is a cloud robotics platform that provides a relaxed optimization algorithm for fine-grained cost and performance tradeoffs. 


\subsection{Fine-Grained Cost Analysis}
\label{sec:desing:cost}
\algname can simultaneously benchmark
on multiple cloud machines and dynamically collect the timing for specific ROS2 applications. One can specify potential constraints such as hardware requirement (CPU count, GPU type, and memory size) or cloud machine type specific to the cloud service providers. 
The ROS2 timing is collected from the callback of the ROS2 applications. 
\\
\textbf{ROS2 Latency Collection}.
\algname runs a process that passively collects the request and response ROS topics. It uses the difference between the response message and request message to determine the latency. Alternatively, users can use \algname callback at the start and end of the desired latency collection period that requires three line modification in the application code. 

\textbf{Latency-to-Cost Calculation}.
Once the benchmark condition is fulfilled, such as when the set time has elapsed or the requisite number of latency statistics is collected, \algname collects the data from all the candidates and queries the SkyPilot \cite{286502} interface for the cost of cloud machines.
\algname utilizes the hourly rate information gathered from SkyPilot to calculate the cost of using a machine for a single second. The total cost for a specific ROS2 request is determined by multiplying this per-second cost by the number of seconds required to complete the service.

\subsection{\algname Optimizer}
\label{sec:desing:optimizer}




We set up a relaxed optimization problem of minimizing the latency per request $latency(x)$ and financial cost per request $cost(x)$ with hardware specification $x = \{x_1, x_2\}$, where $x_1$ is the CPU core count and $x_2$ is the memory size in gigabytes (GB). 
The optimization problem can be formulated as minimizing: 
{\small
\begin{align*}
    \alpha \cdot \text{latency}(x) + (1 - \alpha) \cdot \text{cost}(x)
\end{align*}}
where latency and cost are expressed in dimensionless units.

{\small
\textbf{With System Constraints:}
\begin{align*}
    x_1 &\geq \text{Min CPU} & x_1 &\leq \text{Max CPU} \\
    x_2 &\geq \text{Min Memory} & x_2 &\leq \text{Max Memory} \\
    \frac{x_2}{x_1} &\geq \text{Min Memory:CPU} & \frac{x_2}{x_1} &\leq \text{Max Memory:CPU} \\
\end{align*}

\textbf{With User Constraints:}
\begin{align*}
    \text{latency}(x) &\leq \text{Max Latency} \\
    \text{cost}(x) &\leq \text{Max Cost}
\end{align*}
}

Hyperparameter $\alpha \in [0, 1] $ weighs the cost and latency tradeoff of the objective function. Since every user may have their own preference on how time and cost should be weighted for a given task, we divide the problem into two parts: minimizing financial cost and minimizing latency (ROS2 node processing time). $\alpha = 0$ favors tasks requiring precise latency demands such as dynamic robot control while  $\alpha = 1$ favors tasks with more relaxed latency demands such as long-horizon robotic exploration.
CPU count and memory are treated as primary variables since they are continuous. 
However, it is important to note that CPU count and memory are discrete. 
As $x$ must be integer-valued, we relax the mixed integer optimization by allowing $x$ to take on contiguous values. We then explore admissible-nearest-neighbors to the resulting solutions. 
From there, we compare the feasible options and select the solution from the subset that minimizes loss and satisfies constraints.

The GPU type introduces an additional challenge due to its discrete nature. To address this issue, the optimizer fits a custom latency and cost function to each GPU type and solve separate optimization problems. The results from these problems are then compared with each other and the program returns the hardware configurations that meet the constraints while best minimizing the loss function.




Since the user inputs the desired number of benchmark steps, we can use the benchmarking data to create latency and cost functions. Even though the nature of these functions is unknown, we approximate them by fitting multiple regression functions, gausssian processes and neural networks. We choose the best-fit function while also being careful not to pick functions that overfit.



\subsection{Specification Sampler}
The \algname optimizer relies on being able to
determine time and cost as a function of hardware. Since
there is no detailed understanding of how time and cost relate
to hardware and it is infeasible to test every hardware configuration, we employ sky benchmarking to fit functions
for these parameters. Benchmarking leverages multithreading
for parallel processing to significantly speed up the optimizer.
Specifically, we simultaneously sample edge case hardware
configurations and run them for 5 minutes to determine
seconds per benchmark step and dollars per second. Example
values used to evaluate Dex-Net are outlined in Table \ref{tab:opt:sample}.
\label{sec:design:sampler}
\begin{table}
  \centering
  \renewcommand{\arraystretch}{1} 
  \small
  \begin{tabular}{|c|c|c|c|c|c|}
    \hline
    \cellcolor{white} & \cellcolor{white}Min Mem. & \cellcolor{white}... & \cellcolor{white}Avg Mem. & \cellcolor{white}... & \cellcolor{white}Max Mem. \\
    \hline
    \cellcolor{white}Min CPU & \cellcolor{green!20}(2,4) & \cellcolor{red!20} & \cellcolor{green!20} (2,8) & \cellcolor{red!20} & \cellcolor{green!20} (2,16) \\
    \hline
    \cellcolor{white}... & \cellcolor{red!20} & \cellcolor{red!20} & \cellcolor{red!20} & \cellcolor{red!20} & \cellcolor{red!20} \\
    \hline
    \cellcolor{white}Avg CPU & \cellcolor{green!20}(16,32) & \cellcolor{red!20} & \cellcolor{green!20} (16,64) & \cellcolor{red!20} & \cellcolor{green!20} (16,128) \\
    \hline
    \cellcolor{white}... & \cellcolor{red!20} & \cellcolor{red!20} & \cellcolor{red!20} & \cellcolor{red!20} & \cellcolor{red!20} \\
    \hline
    \cellcolor{white}Max CPU & \cellcolor{green!20}(64,128) & \cellcolor{red!20} & \cellcolor{green!20} (64,256) & \cellcolor{red!20} & \cellcolor{green!20} (64, 512) \\
    \hline
  \end{tabular}
  \caption{\textbf{Hardware Configuration Sampling Grid for Online Benchmarking}. We target edge case hardware combinations to account for the spread of data. For that reason, we sample the minimum, average, and maximum available values for memory (GB) and CPU count (shown in green).}
  \label{tab:opt:sample}
\end{table}

\subsection{Cloud Machine Selector}
\label{sec:desing:selector}

\begin{listing}[t]
\begin{minted}{py}
def generate_launch_description():
    fogros2.SkyLaunchDescription([
        # grasp planning algorithm
        Node(
            package="fogros2_examples", 
            executable="grasp_planning_node",
        ),
    ], constraint = {
        latency: 0.01,  # second
        cost: 0.01      # cent per request
    })

    return LaunchDescription([
        # camera and controller node
        Node(
            package="fogros2_examples", 
            executable="grasp_planning_client", 
        ),
    ])
\end{minted}%
\caption{\textbf{\algname{} Launch Script Example.} In this example, the grasp planning node is launched to the cloud with constraints for cost of less than \$0.01 and 0.01 seconds of latency per request. } 
\label{lst:launch}
\end{listing}

With \algname,  users need only input the per-request application latency (\ref{lst:launch}), and per-request cost for the designated  set of ROS2 nodes (i.e. grasp planning, motion planning, etc.) and its corresponding ROS2 setup configuration and launch file. 
The \algname Cloud Machine Selector supports changing the constraints without re-running the benchmark. 
The users can tune their cost and performance tradeoff by directly applying the constraints on the cost and latency models, without re-running the benchmarks. 
When the actual ROS2 applications are deployed, the cloud machines are initialized from the earlier generated models.

When running \algname for the first time or if the price structure changes, it automatically benchmarks and models the offloaded ROS2 application. A user may optionally specify the benchmark step count that the user wants to benchmark in the algorithm. After the benchmark, \algname automatically deallocates the resources to save the benchmark cost. 


\section{Evaluation}

Without modifying the original ROS2 application code, we integrate \algname with three cloud robotics applications: visual SLAM with ORB-SLAM2~\cite{mur2017orb}, grasp planning with Dex-Net~\cite{mahler2017dex}, and motion planning with Motion Planning Templates (MPT)~\cite{ichnowski2019mpt}. The details of the applications can be found in our earlier FogROS papers~\cite{chen2021fogros, ichnowski2022fogros, chen_fogros2-sgc_2023}.
\begin{table}[t]
    \centering
    \footnotesize
    \begin{tabular}{@{}l@{\quad}c@{\quad} c@{\quad}c@{\quad}c@{\quad}c@{\quad}c@{\quad}c@{}}\toprule
                 &
                  vSLAM&
                  Dex-Net &
                  \multicolumn{3}{c}{Motion Planning}
                   \\
                  \cmidrule{4-6} 
         
     Scenarios     & - & - &  Cubicles & Apartment  \\
         \midrule
 
         Full Benchmark & 201.43 & 205.00 & 178.03 & 178.74  \\
         \algname Sampling  & 104.67 & 117.67 & 103.84 & 104.89   \\
         Cost Reduction& 1.92x & 1.74x & 1.71x& 1.70x \\
         \bottomrule
    \end{tabular}
    \caption{\textbf{Total Benchmark Cost (US Cents) of Benchmarking SLAM, Dex-Net, and Motion Planning} The duration excludes setup time and extrapolated due to the hardware differences, and every specification is tested for a duration of at least 5 minutes. \algname Optimizer reduces the total benchmark cost by up to 1.92 times compared to running through the Full Benchmark, which iterates through all the hardware options. We use fr1/xyz dataset from ORB-SLAM2 \cite{openvslam2019}. }
    \label{tab:eval:benchmark}
\end{table}

\textbf{Pareto Frontier Analysis.}
We evaluate the \algname optimizer using a Pareto frontier to represent the trade-offs between conflicting objectives. A solution is said to be more efficient (non-dominated) if there is no other solution that is better in at least one objective without being worse in at least one other objective. In \algname, a cloud configuration selection is on the Pareto frontier if there is no other selection with cheaper cost \emph{and} lower application latency. It is important to note that the algorithm does not optimize the Pareto frontier staircase curve because the frontier outlines the most efficient cloud instances independent of user-input time and cost constraints. However, the optimizer will minimize either time or cost by finding the minimum value while simultaneously pushing the constraint to its user-input threshold to maximize performance. Thus, the goal of the Pareto frontier analysis is to show that for different scenarios, the sampled latency and cost functions developed by the optimizer are very similar to the ground truth model generated by running all available hardware configuration options. 
We evaluate with commonly used and distinct AWS cloud specifications up to 64 CPU cores as the feasible choices for the Pareto frontier, which the \algname optimizer selects from the available set. Table \ref{tab:eval:benchmark} shows the benchmark duration and the cost of the \algname optimizer comparison.

\textbf{\algname Setup} For fitting the cost and latency functions, we perform constant, linear, hyperbolic, power, log or exponential regressions. We chose regression-based models opposed to Gaussian processes or neural networks to mitigate the risk of overfitting. 

\textbf{\algname Evaluation} We assessed the quality of the regression models by calculating the determination coefficient value, represented as $R^{2}$, a statistical measure that quantifies the proportion of the variance in the dependent variable that is predictable from the independent variable. We use determination coefficient to quantify how well a given regression model fits a dataset where 0 indicates poor/no fit while 1 indicates a perfect fit.

\begin{figure}
\centering
\begin{subfigure}[b]{0.75\linewidth}
    \centering
      \caption{\textbf{Visual SLAM} }
    \includegraphics[width=\linewidth]{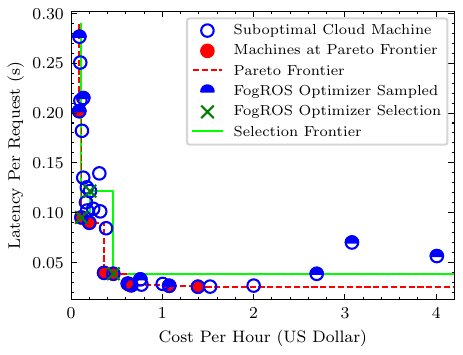}
    \label{fig:pareto:slam}
\end{subfigure}
\begin{subfigure}[b]{0.75\linewidth}
    \centering
        \caption{\textbf{Grasp Planning} }
    \includegraphics[width=\linewidth]{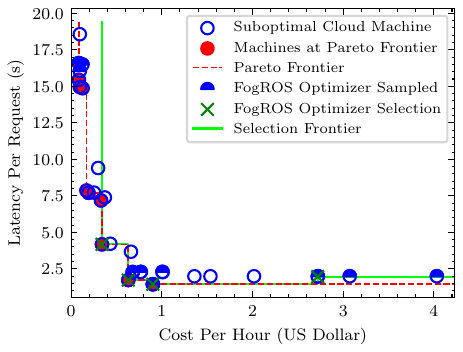}
    \label{fig:pareto:dexnet}
\end{subfigure}
\caption{\textbf{Pareto Frontier Analysis of \algname on Visual SLAM with Orb-SLAM and Grasp Planning with Dex-Net}. We generated a ground truth Pareto frontier plot (red) along with a Pareto frontier made from the hardware configurations sampled by the optimizer (green). In both cases, we can see that both lines follow a similar staircase pattern indicating a strong fit.}
\end{figure}

\subsection{Case Study: Visual SLAM}

For SLAM, the time and cost functions for the T4 GPU case and no GPU case had determination coefficients above 0.95, signifying a strong fit to the data. This strong fit explains why the benchmark-sampled Pareto frontier (depicted in green) closely resembles the ground truth Pareto frontier (depicted in red) in Figure \ref{fig:pareto:slam}. Furthermore, the ground truth machine's frontier consistently deviates from the Sky optimizer's Pareto frontier by no more than 0.0265 seconds. Notably, the optimizer's selection includes two types of c6i instances, characterized by being CPU-only with a memory to CPU ratio of 2 and no GPU. This selection is logical, as the visual SLAM algorithm involved in this process neither makes use of GPUs nor requires excessive memory for computation, rendering any expenditure on them purely inefficient. FogROS\cite{chen2021fogros} used AWS c4.8xlarge for vSLAM, which costs \$1.99 per hour, which suggests a 4.95x cost reduction with the similar latency.

\subsection{Case Study: Grasp Planning with Dex-Net}
For Dex-Net, all time and cost function determination coefficients were above 0.95 for the T4 GPU case and no GPU case, signifying a strong fit to the data. Similar to SLAM, the benchmark-sampled frontier closely matches the shape of the ground truth frontier, as seen in Figure \ref{fig:pareto:dexnet}. The frontiers are equal for costs below \$2.72, and the maximum latency difference  for other costs is 0.286 seconds. However, for Dex-Net, the optimizer selected 2 gd4n instances, characterized by using a T4 GPU. This selection is logical for Dex-Net because it involves deploying a Grasp Quality Convolutional Neural Network (GQ-CNN), which requires GPU for optimal performance.




%


\subsection{Case Study: Motion Planning}

\begin{figure}
     \centering
     \begin{subfigure}[b]{0.73\linewidth}
        \centering
         \caption{\textbf{Cubicle}}
        \includegraphics[width=\linewidth]{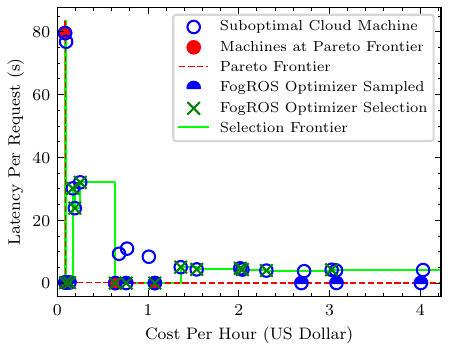}
        \label{fig:mp:cu:lat}
     \end{subfigure}
     \begin{subfigure}[b]{0.73\linewidth}
        \centering
        \caption{\textbf{Apartment}}
        \includegraphics[width=\linewidth]{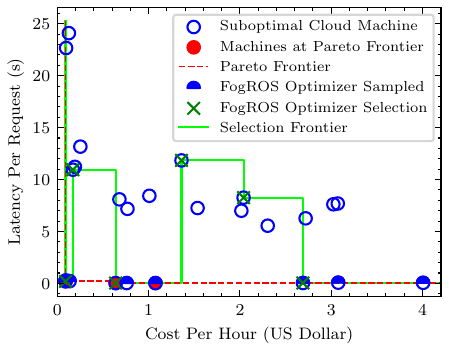}
        \label{fig:mp:ap:lat}
    \end{subfigure}
    \caption{\textbf{Cost-Latency Pareto Frontier Analysis of Motion Planning}. Due to the scholastic nature of the motion planner, the apartment test scenario demonstrates a weak correlation between latency and cloud resource costs. This leads the \algname optimizer to make sub-optimal  selections. }
    \label{fig:mp:cl}
\end{figure}
\begin{figure}
    \begin{subfigure}[b]{0.73\linewidth}
        \centering
        \caption{\textbf{Cubicle}}
        \includegraphics[width=\linewidth]{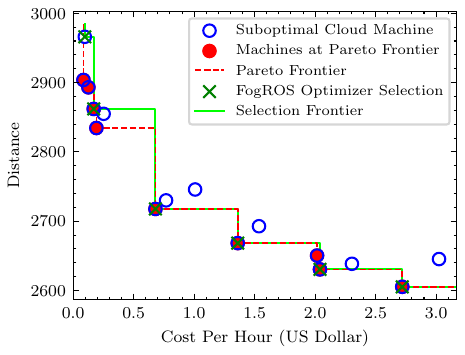}
        \label{fig:mp:cu:cost}
     \end{subfigure}
    \begin{subfigure}[b]{0.73\linewidth}
        \centering
        \caption{\textbf{Apartment}}
        \includegraphics[width=\linewidth]{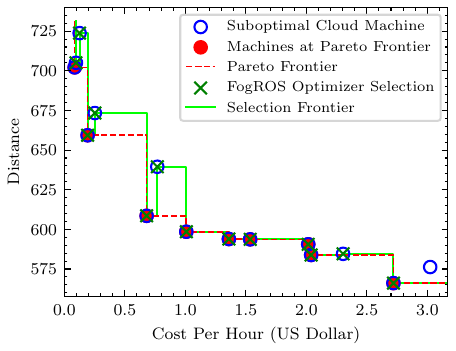}
        \label{fig:mp:ap:cost}
    \end{subfigure}
    \caption{\textbf{Cost-Motion Planner Distance Pareto Frontier Analysis of Motion Planning}. By co-designing the optimization with the motion planner objective path cost, the \algname optimizer can optimize and derive the most cost-effective cloud machine configuration in a majority of the cost and latency constraint combinations.  }
    \label{fig:mp:cd}
\end{figure}

We benchmarked the \algname on two different motion planning scenarios from the Open Motion Planning Library (OMPL)~\cite{OMPL}: cubicles and apartment. OMPL rotates and translates a rigid-body object along a collision-free path from a start pose to a goal pose, thereby posing different computation and memory requirements between our scenarios.

We demonstrate the capability of \algname by evaluating with PRRT*~\cite{ichnowski2014scalable}, an asymptotically-optimal motion planner. It is able to find shorter, more ideal motion plans by using a longer running time or more CPU computational cores.
Figure \ref{fig:mp:cl} shows the cloud computing cost and application latency analysis for this motion planning approach. In the apartment scenario, anomalies with high cost and high latency samples from both the ground-truth dataset and the \algname optimizer lead to inaccurate cost-latency estimates. FogROS2~\cite{ichnowski2022fogros} used a 96-core cloud machine with \$4.99 per hour cost with 38 second latency; \algname achieves the same latency while reducing the cost by 20 times with a cloud machine with \$0.25 per hour.

In Figure \ref{fig:mp:cd}, we use the extensibility of \algname by co-designing  the objective function with the path distance of the motion plan. This co-designed algorithm significantly facilitates the selection of the cost-effective hardware specification by strongly fitting to the Pareto front.

\section{Limitations and Future Work}
\algname provides the latency-cost tradeoff from user's application constraints.
With respect to latency modeling, we consider processing time and in future work will also incorporate bandwidth optimization for data transmission.
\algname considers major hardware specifications, such as CPU cores, GPU, and memory size, which are critical to the majority of the robotics applications. The paper does not consider complex hardware architectural differences that may inherently influence the performance modeling. In future work, we will explore the use of Spot Instances, a special pricing model provided by majority of cloud service providers that can offer up to 90\% price reduction for cloud machines that do not guarantee availability.

\section*{Acknowledgements}
This research was performed at the AUTOLAB at UC Berkeley in affiliation with the Berkeley AI Research (BAIR) Lab. The authors were supported in part by donations from Bosch. The research is also supported by C3.ai Digital Transformation Institute. We thank Simeon Adebola and Karthik Dharmarajan.








\bibliographystyle{IEEEtran}
\bibliography{IEEEabrv,references}

\end{document}